%% file: main.tex
\documentclass[11pt,a4paper]{article}
\usepackage[hyperref]{acl2021}
\usepackage{times}
\usepackage{latexsym}

\usepackage{microtype}

\usepackage{multirow}
\usepackage{enumitem} 
\usepackage{arydshln} 
\usepackage{booktabs} 

\usepackage{pgfplots}

\usepackage{amsmath,amsfonts,amssymb,amsthm, bm}

\aclfinalcopy 

\usepackage[utf8]{inputenc} 
\usepackage[T1]{fontenc}    
\usepackage{hyperref}       
\usepackage{url}            
\usepackage{booktabs}       
\usepackage{amsfonts}       
\usepackage{nicefrac}       
\usepackage{microtype}      
\usepackage{xcolor,colortbl}
\usepackage{graphicx}
\usepackage{multirow}
\usepackage{ifthen}
\usepackage[detect-weight=true, detect-family=true]{siunitx}
\usepackage{bm}
\usepackage{qtree}
\usepackage{tikz}
\usepackage{arydshln}

\usepackage{algorithm}
\usepackage{algorithmic}

\newcommand*\circled[1]{\tikz[baseline=(char.base)]{
            \node[shape=circle,dashed,draw,inner sep=2pt] (char) {#1};}}

\input{utils/math_commands.tex}

\input{utils/macro.tex}

\newcommand*\samethanks[1][\value{footnote}]{\footnotemark[#1]}

\title{Meta-Learning to Compositionally Generalize}

\author{
    Henry Conklin$^{1}$\Thanks{\ Equal contribution.}, \enskip  Bailin Wang$^{1}$\samethanks, \enskip Kenny Smith$^{1}$ \and Ivan Titov$^{1,2}$ \\
    $^1$University of Edinburgh  \quad $^2$University of Amsterdam \\
    \{henry.conklin, bailin.wang, kenny.smith\}@ed.ac.uk, ititov@inf.ed.ac.uk
}

\date{}

\begin{document}
\maketitle
\begin{abstract}
Natural language is compositional; the meaning of a sentence is a function of the meaning of its parts. This property allows humans to create and interpret novel sentences, generalizing robustly outside their prior experience. Neural networks have been shown to struggle with this kind of generalization, in particular performing poorly on tasks designed to assess compositional generalization (i.e. where training and testing distributions differ in ways that would be trivial for a compositional strategy to resolve).
Their poor performance on these tasks may in part be due to the nature of supervised learning which assumes training and testing data to be drawn from the same distribution. We implement a meta-learning augmented version of supervised learning whose objective directly optimizes for out-of-distribution generalization. We construct pairs of tasks for meta-learning by sub-sampling existing training data. Each pair of tasks is constructed to contain relevant examples, as determined by a similarity metric, in an effort to inhibit models from memorizing their input. Experimental results on the COGS and SCAN datasets show that our similarity-driven meta-learning can improve generalization performance.

\end{abstract}

\input{sections/introduction}
\input{sections/method}

\input{sections/experiments}
\input{sections/discussion}
\input{sections/related_work}
\input{sections/conclusion}

\subsection*{ Acknowledgements}
This work was supported in part by the UKRI Centre for Doctoral Training in Natural Language Processing, funded by the UKRI (grant EP/S022481/1) and the University of Edinburgh, School of Informatics and School of Philosophy, Psychology \& Language Sciences. We also acknowledge the financial support of the European Research Council (Titov, ERC StG BroadSem 678254) and the Dutch National Science Foundation (Titov, NWO VIDI 639.022.518).

\bibliographystyle{acl_natbib}
\bibliography{anthology,main}

\appendix
\input{sections/appendix}

\end{document}

%% file: utils/math_commands.tex
\usepackage{amsmath,amsfonts,bm}

\def\eqref#1{equation~\ref{#1}}

\def\1{\bm{1}}

\DeclareMathAlphabet{\mathsfit}{\encodingdefault}{\sfdefault}{m}{sl}
\SetMathAlphabet{\mathsfit}{bold}{\encodingdefault}{\sfdefault}{bx}{n}



%% file: utils/macro.tex
\newcommand{\round}[1]{\num[round-mode=places,round-precision=1]{#1}}

\newcommand{\acc}[2]{\round{#1}&\ifthenelse{\equal{#2}{}}{}{\tiny ${\scriptstyle \pm}$\round{#2}}}

\newcommand{\graycellcolor}{\cellcolor{gray!25}}

\newcommand{\gacc}[2]{\graycellcolor\round{#1}&\graycellcolor\ifthenelse{\equal{#2}{}}{}{\tiny ${\scriptstyle \pm}$\round{#2}}}

\newcommand{\ACC}[2]{\textbf{\round{#1}}&\ifthenelse{\equal{#2}{}}{}{\textbf{\tiny ${\scriptstyle \pm}$\round{#2}}}}

\newcommand{\GACC}[2]{\graycellcolor\textbf{\round{#1}}&\graycellcolor\ifthenelse{\equal{#2}{}}{}{\textbf{\tiny ${\scriptstyle \pm}$\round{#2}}}}

\newcommand{\mc}[2]{\multicolumn{#1}{c}{\textbf{#2}}}

\newcommand{\origtrain}{\mathcal{T}}
\newcommand{\loss}{\mathcal{L}}
\newcommand{\param}{\theta}
\newcommand{\mtrainBatch}{\mathcal{B}_{t}}
\newcommand{\mtestBatch}{\mathcal{B}_{g}}
\newcommand{\sBatch}{\mathcal{B}}
\newcommand{\update}{\text{Update}}

%% file: sections/introduction.tex
\section{Introduction}

Compositionality is the property of human language that allows for the meaning of a sentence to be constructed from the meaning of its parts and the way in which they are combined \citep{cann_formal_1993}. By decomposing phrases into known parts we can generalize to novel sentences despite never having encountered them before. In practice this allows us to produce and interpret a functionally limitless number of sentences given finite means \citep{chomsky_aspects_1965}.

Whether or not neural networks can generalize in this way remains unanswered. Prior work asserts that there exist fundamental differences between cognitive and connectionist architectures that makes compositional generalization by the latter unlikely \citep{fodor_connectionism_1988}. However, recent work has shown these models' capacity for learning some syntactic properties. \citet{hupkes2018visualisation} show how some architectures can handle hierarchy in an algebraic context and generalize in a limited way to unseen depths and lengths. Work looking at the latent representations learned by deep machine translation systems show how these models seem to extract constituency and syntactic class information from data \citep{blevins_deep_2018, belinkov_evaluating_2018}. These results, and the more general fact that neural models perform a variety of NLP tasks with high fidelity \citep[eg.][]{vaswani2017attention, dong-lapata-2016-language}, suggest these models have some sensitivity to syntactic structure and by extension may be able to learn to generalize compositionally.

Recently there have been a number of datasets designed to more formally assess connectionist models' aptitude for compositional generalization \citep{kim_cogs_2020, lake2018generalization, hupkes_compositionality_2019}. These datasets frame the problem of compositional generalization as one of out-of-distribution generalization: the model is trained on one distribution and tested on another which differs in ways that would be trivial for a compositional strategy to resolve. A variety of neural network architectures have shown mixed performance across these tasks, failing to show conclusively that connectionist models are reliably capable of generalizing compositionally \cite{keysers2020measuring, lake2018generalization}. Natural language requires a mixture of memorization and generalization \citep{jiang_characterizing_2020-1}, memorizing exceptions and atomic concepts with which to generalize. Previous work looking at compositional generalization has suggested that models may memorize large spans of sentences multiple words in length \citep{hupkes_compositionality_2019, keysers2020measuring}. This practice may not harm in-domain performance, but if at test time the model encounters a sequence of words it has not encountered before it will be unable to interpret it having not learned the atoms (words) that comprise it. 
\citet{griffiths2020understanding} looks at the role of limitations in the development of human cognitive mechanisms. Humans' finite computational ability and limited memory may be central to the emergence of robust generalization strategies like compositionality. A hard upper-bound on the amount we can memorize may be in part what forces us to generalize as we do. Without the same restriction models may prefer a strategy that memorizes large sections of the input potentially inhibiting their ability to compositionally generalize.

In a way the difficulty of these models to generalize out of distribution is unsurprising: supervised learning assumes that training and testing data are drawn from the same distribution, and therefore does not necessarily favour strategies that are robust out of distribution. Data necessarily under-specifies for the generalizations that produced it. Accordingly for a given dataset there may be a large number of generalization strategies that are compatible with the data, only some of which will perform well outside of training \citep{damour_underspecification_2020}. It seems connectionist models do not reliably extract the strategies from their training data that generalize well outside of the training distribution. Here we focus on an approach that tries to to introduce a bias during training such that the model arrives at a more robust strategy.

To do this we implement a variant of the model agnostic meta-learning algorithm \citep[MAML,][] {finn_model-agnostic_2017}. The approach used here follows \citet{wang2020meta} which implements an objective function that explicitly optimizes for out-of-distribution generalization in line with \citet{li2018maml}. \citet{wang2020meta} creates pairs of tasks for each batch (which here we call meta-train and meta-test) by sub-sampling the existing training data. Each meta-train, meta-test task pair is designed to simulate the divergence between training and testing: meta-train is designed to resemble the training distribution, and meta-test to resemble the test distribution. The training objective then requires that update steps taken on meta-train are also beneficial for meta-test. This serves as a kind of regularizer, inhibiting the model from taking update steps that only benefit meta-train. By manipulating the composition of meta-test we can control the nature of the regularization applied. Unlike other meta-learning methods this is not used for few or zero-shot performance. Instead it acts as a kind of meta-augmented supervised learning, that helps the model to generalize robustly outside of its training distribution.

The approach taken by \citet{wang2020meta} relies on the knowledge of the test setting. While it does not assume access to the test distribution, it assumes access to the family of test distributions, from which the actual test distribution will be drawn. While substantially less restrictive than the standard iid setting, it still poses a problem if we do not know the test distribution, or if the model is evaluated in a way that does not lend itself to being represented by discrete pairs of tasks (i.e. if test and train differ in a variety of distinct ways). Here we propose a more general approach that aims to generate meta-train, meta-test pairs which are populated with similar (rather than divergent) examples in an effort to inhibit the model from memorizing its input. Similarity is determined by a string or tree kernel so that for each meta-train task a corresponding meta-test task is created from examples deemed similar.

By selecting for similar examples we design the meta-test task to include examples with many of the same words as meta-train, but in novel combinations. As our training objective encourages gradient steps that are beneficial for both tasks we expect the model to be less likely to memorize large chunks which are unlikely to occur in both tasks, and therefore generalize more compositionally. This generalizes the approach from \citet{wang2020meta}, by using the meta-test task to apply a bias not-strictly related to the test distribution: the design of the meta-test task allows us to design the bias which it applies. It is worth noting that other recent approaches to this problem have leveraged data augmentation to make the training distribution more representative of the test distribution \citep{andreas-2020-good}. We believe this line of work is orthogonal to ours as it does not focus on getting a model to generalize compositionally, but rather making the task simple enough that compositional generalization is not needed. Our method is model agnostic, and does not require prior knowledge of the target distribution. 

We summarise our contributions as follows:
%
\begin{itemize}[noitemsep,topsep=0pt]

\item We approach the problem of compositional generalization with a meta-learning objective that tries to explicitly reduce input memorization using similarity-driven virtual tasks.


\item We perform experiments on two text-to-semantic compositional datasets: COGS and SCAN.
Our new training objectives lead to significant improvements in accuracy
over a baseline parser trained with conventional supervised
learning.~\footnote{Our implementations are available at \url{https://github.com/berlino/tensor2struct-public}.}

\end{itemize}

%% file: sections/method.tex
\section{Methods}
We introduce the meta-learning augmented approach to supervised learning from \citet{li2018maml,wang2020meta} that explicitly optimizes for out-of-distribution generalization. Central to this approach is the generation of tasks for meta-learning by sub-sampling training data. We introduce three kinds of similarity metrics used to guide the construction of these tasks.

\subsection{Problem Definition}

\paragraph{Compositional Generalization}
  \citet[eg.][]{lake2018generalization,kim_cogs_2020} introduce datasets designed to assess compositional generalization. These datasets are created by generating synthetic data with different distributions for testing and training. The differences between the distributions are trivially resolved by a compositional strategy. At their core these tasks tend to assess three key components of compositional ability: systematicity, productivity, and primitive application. Systematicity allows for the use of known parts in novel combinations as in (a). Productivity enables generalization to longer sequences than those seen in training as in (b). Primitive application allows for a word only seen in isolation during training to be applied compositionally at test time as in (c).

\begin{itemize}
    \item[(a)] The cat gives the dog a gift $\rightarrow$ The dog gives the cat a gift
    \item[(b)] The cat gives the dog a gift $\rightarrow$ The cat gives the dog a gift and the bird a gift
    \item[(c)] made $\rightarrow$ The cat made the dog a gift
\end{itemize}
A compositional grammar like the one that generated the data would be able to resolve these three kinds of generalization easily, and therefore performance on these tasks is taken as an indication of a model's compositional ability.

\paragraph{Conventional Supervised Learning}
The compositional generalization datasets we look at are semantic parsing tasks, mapping between natural language and a formal representation. A usual supervised learning objective for semantic parsing is to minimize the negative log-likelihood of the correct formal representation given a natural language input sentence, i.e. minimising
\vspace{-2mm}
\begin{equation}
    \loss_{\sBatch}(\theta) = - \frac{1}{N} \sum_{i=1}^{N}  \log p_{\param}(y | x)
\end{equation}
where $N$ is the size of batch $\sBatch$, $y$ is a formal representation and $x$ is a natural language sentence. This approach assumes that the training and testing data are independent and identically distributed.

\paragraph{Task Distributions}
Following from \citet{wang2020meta}, we utilize a learning algorithm that can enable a parser to benefit from a distribution of virtual tasks, denoted by $p(\tau)$, where $\tau$ refers to an instance of a virtual compositional generalization task that has its own training and test examples.



\subsection{MAML Training}

\input{figures/maml.tex}


Once we have constructed our pairs of virtual tasks we need a training algorithm that encourages compositional generalization in each.
Like \citet{wang2020meta},
we turn to optimization-based
meta-learning algorithms \cite{finn2017model,li2018maml}
and apply DG-MAML (Domain Generalization with Model-Agnostic Meta-Learning),
a variant of MAML \cite{finn2017model}.
Intuitively, DG-MAML encourages optimization on meta-training examples to have a positive
effect on the meta-test examples as well.

During each learning episode of MAML training we randomly sample a task $\tau$
which consists of a training batch $\mtrainBatch$ and a generalization batch $\mtestBatch$ and conduct optimization in two steps, namely \textit{meta-train}
and \textit{meta-test}.

\paragraph{Meta-Train}
The meta-train task is sampled at random from the training data. The model performs one stochastic gradient descent step on this batch
\begin{equation}
    \theta'\leftarrow \theta - \alpha \nabla_{\theta} \loss_{\mtrainBatch}(\theta)
    \label{eq:meta_train}
\end{equation}
where $\alpha$ is the meta-train learning rate.

\paragraph{Meta-Test}

The fine-tuned parameters $\theta'$ are evaluated on the accompanying generalization task, meta-test, by computing their loss on it
denoted as $\loss_{\mtestBatch}(\theta')$. The final objective for a task $\tau$
is then to jointly optimize the following:
\begin{align}
\begin{split}
     \loss_{\tau}(\theta)&=\loss_{\mtrainBatch}(\theta) + \loss_{\mtestBatch}(\theta') \\
     &=\mathcal{L}_{\mtrainBatch}(\theta) + \mathcal{L}_{\mtestBatch}(\theta - \alpha \nabla_{\theta} \mathcal{L}_{\beta}(\theta) )
    \label{eq:maml_obj}
\end{split}
\end{align}
The objective now becomes to reduce the joint loss of both the meta-train and meta-test tasks. Optimizing in this way ensures that updates on meta-train are also beneficial to meta-test. The loss on meta-test acts as a constraint on the loss from meta-train. This is unlike traditional supervised learning ($\loss_{\tau}(\theta)=\loss_{\mtrainBatch}(\theta) + \loss_{\mtestBatch}(\theta)$) where the loss on one batch does not constrain the loss on another.

With a random $\mtrainBatch$ and $\mtestBatch$, the joint loss function can be seen as a kind of generic regularizer, ensuring that update steps are not overly beneficial to meta-train alone. By constructing $\mtrainBatch$ and $\mtestBatch$ in ways which we expect to be relevant to compositionality, we aim to allow the MAML algorithm to apply specialized regularization during training. Here we design meta-test to be similar to the meta-train task because we believe this highlights the systematicity generalization that is key to compositional ability: selecting for examples comprised of the same atoms but in different arrangements. In constraining each update step with respect to meta-train by performance on similar examples in meta-test we expect the model to dis-prefer a strategy that does not also work for meta-test like memorization of whole phrases or large sections of the input.


\input{tables/methods_kernel_examples_1}
\input{figures/alt_tree_kernel_examples}

\subsection{Similarity Metrics}

Ideally, the design of virtual tasks should reflect specific generalization cases for each dataset. However, in practice this requires some prior knowledge of the distribution to which the model will be expected to generalize, which is not always available.
Instead we aim to naively structure the virtual tasks to resemble each other.
To do this we use a number of similarity measures intended to help select examples which highlight the systematicity of natural language.

Inspired by kernel density estimation~\cite{parzen1962estimation}, we define a relevance distribution for each example:
\begin{equation}
   \tilde p ( x', y' | x, y )  \propto \exp \big ( k([x, y], [x', y']) / \eta \big )
   \label{eq:rel_dist}
\end{equation}
where $k$ is the similarity function,
$[x, y]$ is a training example, $\eta$
is a temperature that controls the sharpness of the distribution.
Based on our extended interpretation of relevance, a high $\tilde p$ implies that $[x, y]$ is systematically relevant to $[x', y']$ - containing many of the same atoms but in a novel combination. We look at three similarity metrics to guide subsampling existing training data into meta-test tasks proportional to each example's $\tilde p$.

\paragraph{Levenshtein Distance}
First, we consider Levenshtein distance, a
kind of edit distance widely used to measure the dissimilarity between strings. We compute the negative Levenshtein distance at the word-level between natural language sentences of two examples:
\begin{equation}
    k([x, y], [x', y']) = -1 * \text{LevDistance}(x, x')
\end{equation}
where LevDistance returns the number of edit operations required to transform $x$ into $x'$. See Table~\ref{table:kernel-examples} for examples. 

Another family of 
 similarity metrics for discrete structures are convolution kernels~\cite{haussler1999convolution}.

\paragraph{String-Kernel Similarity}
We use the string subsequence kernel~\cite{lodhi2002text}: %
\begin{equation}
    k([x, y], [x', y']) = \text{SSK}(x, x')
\end{equation}
where SSK computes the number of common subsequences between natural language sentences at the word-level.
See Table~\ref{table:kernel-examples} for examples.~\footnote{We use the normalized convolution kernels in this work, i.e., $k'(x_1, x_2)=k(x_1, x_2)/ \sqrt{k(x_1, x_1) k(x_2, x_2)}$}

\paragraph{Tree-Kernel Similarity}
In semantic parsing, the formal representation $y$ usually has a known grammar which can be used to
represent it as a tree structure.
In light of this we use tree convolution kernels to compute similarity between examples:~\footnote{
Alternatively, we can use tree edit-distance~\cite{zhang1989simple}.}
%
\begin{equation}
    k([x, y], [x', y']) =  \text{TreeKernel}(y, y')
\end{equation}
where the TreeKernel function is a convolution kernel~\cite{collins2001convolution} applied to trees.
Here we consider a particular case
where $y$ is represented as a dependency structure,
as shown in Figure \ref{figure:partial-trees}.
We use the partial tree kernel \citep{moschitti2006efficient} which is designed for application to dependency trees.
%
For a given dependency tree partial tree kernels generate a series of all possible partial trees: any set of one or more connected nodes. Given two trees the kernel returns the number of partial trees they have in common, interpreted as a similarity score. Compared with string-based similarity, this kernel prefers sentences that share common syntactic sub-structures, some of which are not assigned high scores in string-based similarity metrics, as shown in Table \ref{table:kernel-examples}.

Though tree-structured formal representations are more informative in obtaining relevance, not all logical forms can be represented as tree structures.
In SCAN~\cite{lake2018generalization} $y$ are action sequences without given grammars. As we will show in the experiments, string-based similarity metrics have a broader scope of applications but are less effective than tree kernels in cases where $y$ can be tree-structured.

\paragraph{Sampling for Meta-Test}
Using our kernels we compute the relevance distribution in Eq~\ref{eq:rel_dist} to construct virtual tasks for MAML training.
We show the resulting procedure in Algorithm~\ref{algo:maml}.
In order to construct a virtual task $\tau$,
a meta-train batch is first sampled at random from the training data (line \ref{algo:line_sample_mtrain}), then the accompanying meta-test batch is created by sampling examples similar to those in meta-train (line \ref{algo:line_sample_mtest}).
%

We use \emph{Lev-MAML, Str-MAML and Tree-MAML} to denote the meta-training using Levenshtein distance, string-kernel and tree-kernel similarity, respectively.

%% file: figures/maml.tex
\begin{algorithm}[t]
\caption{MAML Training Algorithm}
\label{algo:maml}
\begin{algorithmic}[1]
  \REQUIRE Original training set $\origtrain$
  \REQUIRE Learning rate $\alpha$, Batch size $N$
  \FOR{step $\gets 1$ \TO $T$}
    \STATE Sample a random batch from $\origtrain$ as a virtual training set $\mtrainBatch$  \label{algo:line_sample_mtrain}  \\ 
    \STATE Initialize an empty generalization set $\mtestBatch$
    \FOR{$i$ $\gets 1$ \TO $N$}
        \STATE Sample an example from $\tilde p( \cdot \ | \ \mtrainBatch[i])$
        \label{algo:line_sample_mtest}  \\ 
        \STATE Add it to $\mtestBatch$
    \ENDFOR
    \STATE Construct a virtual task $\tau := (\mtrainBatch, \mtestBatch)$
    \STATE Meta-train update: \label{algo:line_meta_train}  \\ 
    \quad $\param'\leftarrow \param - \alpha \nabla_{\param} \loss_{\mtrainBatch} (\param)$ \\ 
    \STATE Compute meta-test objective: \\ \quad $ \loss_{\tau}(\param) = \loss_{\mtrainBatch}  (\param) + \loss_{\mtestBatch}(\param')$ \\
    \STATE Final Update: \label{algo:line_meta_update} \\ 
    \quad  $\param \leftarrow  \update(\param, \nabla_{\param} \loss_{\tau}(\param))$
   \ENDFOR
\end{algorithmic}
\end{algorithm}

%% file: tables/methods_kernel_examples_1.tex
\begin{table}[t!]
  \centering
    \resizebox{\columnwidth}{!}{ 
      \begin{tabular}{lr}
    \multicolumn{2}{c}{\textbf{Source Example}: The girl changed a sandwich beside the table .} \\
      \\
      \toprule
      \emph{Neighbours using Tree Kernel} & Similarity\\
      \hline
A sandwich changed . & 0.55 \\
The girl changed . & 0.55 \\
The block was changed by the girl . & 0.39 \\
The girl changed the cake . & 0.39 \\
change & 0.32 \\
       \\
        \emph{Neighbours using String Kernel}\\
       \hline
The girl rolled a drink beside the table . & 0.35 \\
The girl liked a dealer beside the table . & 0.35 \\
The girl cleaned a teacher beside the table . & 0.35 \\
The girl froze a bear beside the table . & 0.35 \\
The girl grew a pencil beside the table . & 0.35 \\
      \\
      \emph{Neighbours using LevDistance} & \\
      \hline
The girl rolled a drink beside the table . & -2.00 \\
The girl liked a dealer beside the table . & -2.00 \\
The girl cleaned a teacher beside the table . & -2.00 \\
The girl froze a bear beside the table . & -2.00 \\
The girl grew a pencil beside the table . & -2.00 \\
      \bottomrule
      \end{tabular}
  }
  \caption{Top scoring examples according to the tree kernel, string kernel and Levenshtein distance for the sentence `The girl changed a sandwich beside the table~.' and accompanying scores.}
  \label{table:kernel-examples}
\vspace{-3mm}
  \end{table}

%% file: figures/alt_tree_kernel_examples.tex
\begin{figure}[t]
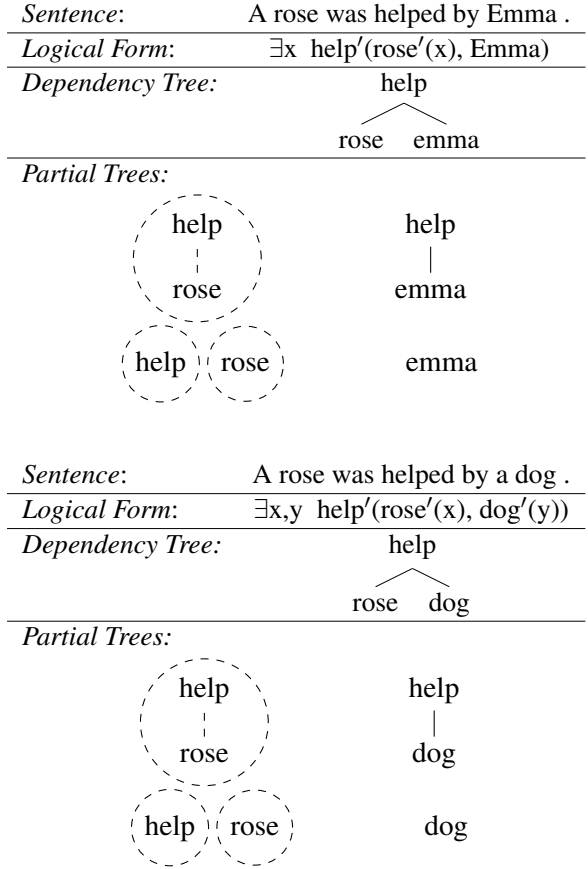

\centering

\resizebox{\columnwidth}{!}{%
\begin{tabular}{lc}
    \emph{Sentence}: & A rose was helped by Emma . \\
    \hline
\emph{Logical Form}: & 
$\exists $x \ help$'$(rose$'$(x), Emma) \\
\hline
    \emph{Dependency Tree:} &
\Tree [.help rose emma ] \\
\hline
\emph{Partial Trees:} &
\end{tabular}
}
\circled{\Tree [.help rose ]}
\Tree [.help emma ] \\
\circled{\Tree [.help ]}
\circled{\Tree [.rose ]}
\Tree [.emma ] \\[1.5\baselineskip]

\resizebox{\columnwidth}{!}{%
\begin{tabular}{lc}
\emph{Sentence}: & A rose was helped by a dog .  \\
\hline
\emph{Logical Form}: & 
$\exists$x,y \ help$'$(rose$'$(x), dog$'$(y)) \\
\hline
\emph{Dependency Tree:} &
\Tree [.help rose dog ] \\
\hline
\emph{Partial Trees:} & 
\end{tabular}
}

\circled{\Tree [.help rose ]}
\Tree [.help dog ] \\
\circled{\Tree [.help ] } \circled{\Tree [.rose ] } \Tree [.dog ] 
\caption{The dependency-tree forms for the logical forms of two sentences.  Shown below each tree are its partial trees. As there are three partial trees shared by the examples their un-normalized tree kernel score is 3.}
\label{figure:partial-trees}
\vspace{-3mm}
\end{figure}

%% file: sections/experiments.tex
\section{Experiments}


\subsection{Datasets and Splits}
We evaluate our methods on the following semantic parsing benchmarks
that target compositional generalization.

\paragraph{SCAN} contains a set of natural language commands and their corresponding 
action sequences~\cite{lake2018generalization}. We use the Maximum Compound Divergence (MCD) 
splits~\cite{keysers2020measuring}, which are created based on the principle of 
maximizing the divergence between the compound (e.g., patterns of 2 or more action sequences) distributions of
the training and test tests. We apply 
Lev-MAML and Str-MAML to SCAN where similarity measures are applied
to the natural language commands. Tree-MAML (which uses a tree kernel) is not applied as the action sequences do not have an underlying dependency tree-structure. 

\paragraph{COGS} contains a diverse set of natural language sentences paired with logical 
forms based on lambda calculus~\cite{kim_cogs_2020}. Compared with SCAN, it covers various systematic linguistic abstractions 
(e.g., passive to active) including examples of lexical and structural generalization, and thus better reflects the compositionality of natural language. 
In addition to the standard splits of Train/Dev/Test,
COGS provides a generalization (Gen) set drawn from a different distribution that specifically assesses compositional generalization. We apply Lev-MAML, Str-MAML and Tree-MAML to COGS; Lev-MAML and Str-MAML make use of
the natural language sentences while Tree-MAML uses the dependency structures reconstructed from the logical forms.

\subsection{Baselines}

In general, our method is model-agnostic and can be coupled with any semantic parser to improve its compositional generalization. Additionally Lev-MAML, and Str-MAML are dataset agnostic provided the dataset has a natural language input.
In this work, we apply our methods on two widely used sequence-to-sequences models.~\footnote{
Details of implementations and hyperparameters 
can be found in the Appendix.}

\paragraph{LSTM-based Seq2Seq} has been the backbone of many neural semantic parsers~\cite{dong-lapata-2016-language,jia-liang-2016-data}.
It utilizes LSTM~\cite{hochreiter1997long} and attention~\cite{bahdanau2014neural} under
an encoder-decoder~\cite{sutskever2014sequence} framework.

\paragraph{Transformer-based Seq2Seq} also follows the encoder-decoder framework, but 
it uses Transformers~\cite{vaswani2017attention}
to replace the LSTM for encoding and decoding. It has proved successful in many NLP
tasks e.g., machine translation. Recently,
it has been adapted for semantic parsing~\cite{wang-etal-2020-rat} with superior performance.

We try to see whether our  MAML training
can improve the compositional generalization of contemporary semantic parsers, compared with standard supervised learning. Moreover, we include a meta-baseline, referred to as Uni-MAML, 
that constructs meta-train and meta-test splits by uniformly sampling training examples. 
By comparing with this meta-baseline, we show the effect of similarity-driven construction of meta-learning splits.
Note that we do not focus on making comparisons with other methods that feature 
specialized architectures for SCAN datasets (see Section~\ref{sec:related_work}), as these methods do not generalize well to more complex datasets~\cite{furrer2020compositional}.

\paragraph{GECA} We additionally apply the good enough compositional augmentation (GECA) method laid out in \citet{andreas-2020-good} to the SCAN MCD splits. Data augmentation of this kind tries to make the training distribution more representative of the test distribution. This approach is distinct from ours which focuses on the training objective, but the two can be combined with better overall performance as we will show. Specifically, we show the results of GECA applied to the MCD splits as well as GECA combined with our Lev-MAML variant. Note that we elect not to apply GECA to COGS, as the time and space complexity~\footnote{See the original paper for details.} of GECA proves very costly for COGS in our preliminary experiments.

\subsection{Construction of Virtual Tasks}

The similarity-driven sampling distribution $\tilde p$ in Eq~\ref{eq:rel_dist} requires computing 
the similarity between every pair of training examples, which can be very expensive depending on 
the size of of the dataset. As the sampling distributions are fixed during training, we compute 
and cache them beforehand. However, they take an excess of disk space to store as essentially 
we need to store an $N \times N$ matrix where $N$ is the number of training examples. 
To allow efficient storage and sampling, we use the following approximation.
First, we found that usually each example only has a small set of neighbours that are relevant to it.~\footnote{
For example, in COGS, each example only retrieves 3.6\% of the whole training set as its neighbours 
(i.e., have non-zero tree-kernel similarity) on average.}
Motivated by this observation, we only store the top 1000 relevant neighbours for each example sorted by similarity, 
and use it to construct the sampling distribution denoted as $\tilde p_{\text{top}1000}$. 
To allow examples out of top 1000 being sampled, we use a linear interpolation 
between $\tilde p_{\text{top}1000}$ and a uniform distribution. 
Specifically, we end up using the following sampling distribution:
\begin{equation*}
  \tilde p(x', y' | x, y) = \lambda \ \tilde  p_{\text{top}1000} ( x', y' | x, y ) + (1 - \lambda) \frac{1}{N}
\end{equation*}
where $\tilde p_{\text{top}1000}$ assigns 0 probability to out-of top 1000 examples, $N$ is the number of training examples, 
and $\lambda$ is a hyperparameter for interpolation. 
In practice, we set $\lambda$ to $0.5$ in all experiments.
To sample from this distribution, we first decide whether the sample is in the top 1000 by sampling 
from a Bernoulli distribution parameterized by $\lambda$. 
If it is, we use $\tilde p_{\text{top1000}}$ to do the sampling; 
otherwise, we uniformly sample an example from the training set. 

\input{tables/scan_results_seed0-14-conf.tex}

\subsection{Development Set}
Many tasks that assess out-of-distribution (O.O.D.) generalization (e.g. COGS) do not have an O.O.D. Dev set that is representative of the generalization distribution. This is desirable as a parser in principle should never have knowledge of the Gen set during training. In practice though the lack of an O.O.D. Dev set makes model selection extremely difficult and not reproducible.~\footnote{
We elaborate on this issue in the Appendix.}
In this work, we propose the following strategy to alleviate this issue: 1) we sample a small subset from the Gen set, denoted as `Gen Dev' for tuning meta-learning hyperparmeters, 2) we use two disjoint sets of random seeds for development and testing respectively, i.e., retraining the selected models from scratch before applying them to the final test set. In this way, we make sure that our tuning is not exploiting the models resulting from specific random seeds: we do not perform random seed tuning.
At no point are any of our models trained on the Gen Dev set.

\input{tables/cogs_results_seed0-14.tex}

\subsection{Main Results}
On SCAN, as shown in Table~\ref{table:mcd-scan-results}, Lev-MAML substantially helps both base parsers achieve better performance across three different splits constructed according to the MCD principle.~\footnote{Our base parsers also perform much better than previous methods, likely due to the choice of hyperparameters.} 
Though our models do not utilize pre-training such as T5~\cite{raffel2019exploring}, our best model (Lev-MAML + LSTM) still outperforms T5 based models significantly in MCD1 and MCD2. We show that GECA is also effective for MCD splits (especially in MCD1). More importantly, augmenting GECA with Lev-MAML further boosts the performance substantially in MCD1 and MCD2, signifying that our MAML training is complementary to GECA to some degree.

Table~\ref{table:cogs-results} shows our results on COGS. Tree-MAML boosts the performance of both LSTM and Transformer base parsers by a large margin: 6.5\% and 8.1\% respectively in average accuracy. 
Moreover, Tree-MAML is consistently better than other MAML variants, showing the effectiveness of exploiting tree structures of formal representation to construct virtual tasks.
\footnote{The improvement of all of our MAML variants applied to the Transformer are significant (p < 0.03) compared to the baseline, of our methods applied to LSTMs, Tree-MAML is significant (p < 0.01) compared to the baseline.}


\input{tables/cogs_gen_cases.tex}

%% file: tables/scan_results_seed0-14-conf.tex
\begin{table}[t]
\centering
\setlength{\tabcolsep}{3pt}
\resizebox{\columnwidth}{!}{
\begin{tabular}{lrlrlrl}
\textbf{Model}             &  \mc{2}{MCD1} & \mc{2}{MCD2} & \mc{2}{MCD3}  \\
\toprule
LSTM                        & \acc{4.7}{2.2}   & \acc{7.3}{2.1} & \acc{1.8}{0.7} \\
Transformer                  & \acc{0.4}{0.4}   & \acc{1.8}{0.4} & \acc{0.5}{0.1} \\
T5-base                        & \acc{26.2}{1.7}  & \acc{7.9}{1.6} & \acc{12.1}{0.1}\\
T5-11B                         & \acc{7.9}{}      & \acc{2.4}{}    & \ACC{16.8}{}\\
\hline
LSTM &  \gacc{27.4}{8.2} & \gacc{31.0}{0.4} & \gacc{9.6}{3.7}  \\
\ \emph{w.} Uni-MAML &  \gacc{44.8}{5.4} & \gacc{31.9}{3.4} & \gacc{10.0}{1.4}  \\
\ \emph{w.} Lev-MAML &  \GACC{47.6}{2.3} & \GACC{35.2}{3.9} & \gacc{11.4}{3.0}  \\
\ \emph{w.} Str-MAML &  \gacc{42.2}{2.6} & \gacc{33.6}{4.3} & \gacc{11.4}{2.2}  \\
\hline
Transformer &  \gacc{2.6}{0.8} & \gacc{3.1}{1.0} & \gacc{2.3}{1.3}  \\
\ \emph{w.} Uni-MAML &  \gacc{2.8}{0.7} & \gacc{3.2}{1.0} & \gacc{3.2}{1.6}  \\
\ \emph{w.} Lev-MAML &  \gacc{4.7}{1.8} & \gacc{6.7}{1.4} & \gacc{6.5}{1.2}  \\
\ \emph{w.} Str-MAML &  \gacc{2.8}{0.6} & \gacc{5.6}{1.6} & \gacc{6.7}{1.4}  \\
\hline
GECA + LSTM &  \gacc{51.5}{4.4} & \gacc{30.4}{4.8} & \gacc{12.0}{6.8}  \\
\ \emph{w.} Lev-MAML &  \GACC{58.9}{6.4} & \GACC{34.5}{2.5} & \gacc{12.3}{4.9}  \\
\bottomrule
\end{tabular}
}
\setlength{\tabcolsep}{6pt}
\caption{Main results on SCAN MCD splits.  
We show the mean and variance (95\% confidence interval) of 10 runs. 
Cells with a grey background are results obtained in this paper, whereas cells 
with a white background are from~\citet{furrer2020compositional}. }
\label{table:mcd-scan-results}
\end{table}

%% file: tables/cogs_results_seed0-14.tex
\begin{table}[t]
\centering
\setlength{\tabcolsep}{3pt}
\resizebox{\columnwidth}{!}{
    \begin{tabular}{lrlrlrl}
    \textbf{Model}             &  \mc{2}{Gen Dev} & \mc{2}{Test} & \mc{2}{Gen}  \\
    \toprule
    LSTM                        & - &    & \acc{99}{} & \acc{16}{8} \\
    Transformer                  & - &   & \acc{96}{} & \acc{35}{6} \\
    \hline
    LSTM &  \gacc{30.3}{7.3} & \gacc{99.7}{} & \gacc{34.5}{4.5}  \\
    \  \emph{w.} Uni-MAML & \gacc{36.1}{6.7}  & \gacc{99.7}{} & \gacc{36.4}{3.6} \\
    \ \emph{w.} Lev-MAML & \gacc{35.6}{5.3} & \gacc{99.7}{} & \gacc{36.4}{5.2}  \\
    \ \emph{w.} Str-MAML &  \gacc{36.3}{4.2} & \gacc{99.7}{} & \gacc{36.8}{3.5}  \\
    \  \emph{w.} Tree-MAML  & \GACC{41.2}{2.8}  & \gacc{99.7}{} & \GACC{41.0}{4.9} \\
    \hline
    Transformer &  \gacc{54.7}{4.0} & \gacc{99.5}{} & \gacc{58.6}{3.7}  \\
    \  \emph{w.} Uni-MAML & \gacc{60.9}{2.8}  & \gacc{99.6}{} & \gacc{64.4}{4.0} \\
    \  \emph{w.} Lev-MAML & \gacc{62.7}{3.8}  & \gacc{99.7}{} & \gacc{64.9}{6.3} \\
\ \emph{w.} Str-MAML &  \gacc{62.3}{3.0} & \gacc{99.6}{} & \gacc{64.8}{5.5}  \\
    \  \emph{w.} Tree-MAML & \GACC{64.1}{3.2}  & \gacc{99.6}{} & \GACC{66.7}{4.4} \\
    \bottomrule
    \end{tabular}
}
\setlength{\tabcolsep}{6pt}
\caption{Main results on the COGS dataset.  
We show the mean and variance (standard deviation) of 10 runs. 
Cells with a grey background are results obtained in this paper, whereas cells 
with a white background are from~\citet{kim_cogs_2020}.}
\label{table:cogs-results}
\end{table}

%% file: tables/cogs_gen_cases.tex
\pgfplotsset{height=2.5cm, width=6cm, compat=1.9}
\pgfplotsset{xtick style={draw=none}}
\pgfplotsset{ytick style={draw=none}}

\begin{table*}[h]
    \centering
    \resizebox{2\columnwidth}{!}{\begin{tabular}{p{4.2cm}p{4.5cm}p{4.5cm}c}
    \toprule
         Case &  Training & Generalization &  Accuracy Distribution  \\ \midrule
        Primitive noun $\rightarrow$ Subject \newline (common noun) & \textbf{shark} & A \textbf{shark} examined the child. &     
             \raisebox{-0.7\totalheight}{\input{figures/gen_case_prim_sub_common.tex}}  \\ 
        Primitive noun $\rightarrow$ Subject \newline (proper noun) & \textbf{Paula} & \textbf{Paula} sketched William. &
             \raisebox{-0.7\totalheight}{\input{figures/gen_case_prim_sub_proper.tex}}  \\ 
        Primitive noun $\rightarrow$ Object (common noun) & \textbf{shark} & A chief heard the \textbf{shark}.  &
             \raisebox{-0.7\totalheight}{\input{figures/gen_case_prim_obj_common.tex}}  \\ 
        Primitive noun $\rightarrow$ Object \newline (proper noun) & \textbf{Paula} & The child helped \textbf{Paula}. &
             \raisebox{-0.7\totalheight}{\input{figures/gen_case_prim_obj_proper.tex}}  \\ 
       \bottomrule
    \end{tabular}}
    \caption{Accuracy on COGS by generalization case. Each dot represents a single run of the model.}
    \label{table:case-results}
\vspace{-3mm}
\end{table*}

%% file: figures/gen_case_prim_sub_common.tex
\begin{tikzpicture}
    \begin{axis}[ 
      symbolic y coords={Baseline, Tree-MAML},
      ytick={Baseline, Tree-MAML},
    ]
      \addplot[only marks, color=blue, scatter src=explicit symbolic, mark=*, mark size=1.9pt, domain=0.0:1.0] 
      coordinates {
(0.921,Baseline) (0.242,Baseline) (0.292,Baseline) (0.985,Baseline) (0.946,Baseline) (0.11,Baseline) (0.508,Baseline) (0.967,Baseline) (0.885,Baseline) (0.851,Baseline)
(0.949,Tree-MAML) (0.977,Tree-MAML) (0.916,Tree-MAML) (0.995,Tree-MAML) (0.829,Tree-MAML) (0.876,Tree-MAML) (0.726,Tree-MAML) (0.963,Tree-MAML) (0.812,Tree-MAML) (0.922,Tree-MAML)
      }; 
    \end{axis}
\end{tikzpicture}

%% file: figures/gen_case_prim_sub_proper.tex
\begin{tikzpicture}
    \begin{axis}[ 
      symbolic y coords={Baseline, Tree-MAML},
      ytick={Baseline, Tree-MAML},
    ]
      \addplot[only marks, color=blue, scatter src=explicit symbolic, mark=*, mark size=1.9pt, domain=0.0:1.0] 
      coordinates {
(0.901,Baseline) (0.411,Baseline) (0.708,Baseline) (0.667,Baseline) (0.873,Baseline) (0.939,Baseline) (0.386,Baseline) (0.316,Baseline) (0.729,Baseline) (0.892,Baseline)
(0.806,Tree-MAML) (0.354,Tree-MAML) (0.887,Tree-MAML) (0.835,Tree-MAML) (0.895,Tree-MAML) (0.844,Tree-MAML) (0.766,Tree-MAML) (0.362,Tree-MAML) (0.705,Tree-MAML) (0.958,Tree-MAML)
      }; 
    \end{axis}
\end{tikzpicture}

%% file: figures/gen_case_prim_obj_common.tex
\begin{tikzpicture}
    \begin{axis}[ 
      symbolic y coords={Baseline, Tree-MAML},
      ytick={Baseline, Tree-MAML},
    ]
      \addplot[only marks, color=blue, scatter src=explicit symbolic, mark=*, mark size=1.9pt, domain=0.0:1.0] 
      coordinates {
(0.017,Baseline) (0.069,Baseline) (0.367,Baseline) (0.001,Baseline) (0.0,Baseline) (0.531,Baseline) (0.0,Baseline) (0.001,Baseline) (0.029,Baseline) (0.492,Baseline)
(0.048,Tree-MAML) (0.0,Tree-MAML) (0.512,Tree-MAML) (0.522,Tree-MAML) (0.23,Tree-MAML) (0.185,Tree-MAML) (0.059,Tree-MAML) (0.0,Tree-MAML) (0.422,Tree-MAML) (0.347,Tree-MAML)
      }; 
    \end{axis}
\end{tikzpicture}

%% file: figures/gen_case_prim_obj_proper.tex
\begin{tikzpicture}
    \begin{axis}[ 
      symbolic y coords={Baseline, Tree-MAML},
      ytick={Baseline, Tree-MAML},
    ]
      \addplot[only marks, color=blue, scatter src=explicit symbolic, mark=*, mark size=1.9pt, domain=0.0:1.0] 
      coordinates {
(0.088,Baseline) (0.034,Baseline) (0.17,Baseline) (0.886,Baseline) (0.647,Baseline) (0.028,Baseline) (0.149,Baseline) (0.706,Baseline) (0.747,Baseline) (0.416,Baseline)
(0.66,Tree-MAML) (0.775,Tree-MAML) (0.797,Tree-MAML) (0.965,Tree-MAML) (0.491,Tree-MAML) (0.445,Tree-MAML) (0.239,Tree-MAML) (0.647,Tree-MAML) (0.943,Tree-MAML) (0.48,Tree-MAML)
      }; 
    \end{axis}
\end{tikzpicture}

%% file: sections/discussion.tex
\section{Discussion}

\subsection{SCAN Discussion}

The application of our string-similarity driven meta-learning approaches to the SCAN dataset improved the performance of the LSTM baseline parser. Our results are reported on three splits of the dataset generated according to the maximum compound divergence (MCD) principle. 
We report results on the only MCD tasks for SCAN as these tasks explicitly focus on the systematicity of language. As such they assess a model's ability to extract sufficiently atomic concepts from its input, such that it can still recognize those concepts in a new context (i.e. as part of a different compound). To succeed here a model must learn atoms from the training data and apply them compositionally at test time. The improvement in performance our approach achieves on this task suggests that it does disincentivise the model from memorizing large sections - or entire compounds - from its input. 

GECA applied to the SCAN MCD splits does improve performance of the baseline, however not to the same extent as when applied to other SCAN tasks in \citet{andreas-2020-good}. GECA's improvement is comparable to our meta-learning method, despite the fact that our method does not leverage any data augmentation. This means that our method achieves high performance by generalizing robustly outside of its training distribution, rather than by making its training data more representative of the test distribution. The application of our Lev-MAML approach to GECA-augmented data results in further improvements in performance, suggesting that these approaches aid the model in distinct yet complementary ways.

\subsection{COGS Discussion}

All variants of our meta-learning approach improved both the LSTM and Transformer baseline parsers' performance on the COGS dataset. The Tree-MAML method outperforms the Lev-MAML, Str-MAML, and Uni-MAML versions. The only difference between these methods is the similarity metric used, and so differences in performance must be driven by what each metric selects for. For further analysis of the metrics refer to the appendix.

The strong performance of the Uni-MAML variant highlights the usefulness of our approach generally in improving models' generalization performance. Even without a specially designed meta-test task this approach substantially improves on the baseline Transformer model. We see this as evidence that this kind of meta-augmented supervised learning acts as a robust regularizer particularly for tasks requiring out of distribution generalization. 

Although the Uni-MAML, Lev-MAML, and Str-MAML versions perform similarly overall on the COGS dataset they may select for different generalization strategies. The COGS generalization set is comprised of 21 sub-tasks which can be used to better understand the ways in which a model is generalizing (refer to Table~\ref{table:case-results} for examples of subtask performance). Despite having very similar overall performance Uni-MAML and Str-MAML perform distinctly on individual COGS tasks - with their performance appearing to diverge on a number of of them. This would suggest that the design of the meta-test task may have a substantive impact on the kind of generalization strategy that emerges in the model. For further analysis of COGS sub-task performance see the appendix.

Our approaches' strong results on both of these datasets suggest that it aids compositional generalization generally. However it is worth nothing that both datasets shown here are synthetic, and although COGS endeavours to be similar to natural data, the application of our methods outside of synthetic datasets is important future work. 

%% file: sections/related_work.tex
\section{Related Work}
\label{sec:related_work}

\paragraph{Compositional Generalization}

A large body of work on compositional generalization provide models with strong compositional bias,
such as specialized neural architectures~\cite{li2019compositional,russin2019compositional,gordon2019permutation}, or
grammar-based models that accommodate alignments between natural language utterances and programs~\cite{shaw2020compositional,herzig2020span}.
Another line of work  utilizes data augmentation via fixed rules~\cite{andreas-2020-good} 
or a learned network~\cite{akyurek2020learning} in an effort to transform the out-of-distribution compositional generalization task into an in-distribution one. Our work follows an orthogonal direction, 
injecting compositional bias using a specialized training algorithm.
A related area of research looks at the emergence of compositional languages, often showing that languages which seem to lack natural-language like compositional structure may still be able to generalize to novel concepts \citep{kottur2017natural, chaabouni-etal-2020-compositionality}. This may help to explain the ways in which models can generalize robustly on in-distribution data unseen during training while still struggling on tasks specifically targeting compositionality.

\paragraph{Meta-Learning for NLP}
Meta-learning methods~\cite{vinyals2016matching,ravi2016optimization, finn2017model}
that are widely used for few-shot learning, have been adapted 
for NLP applications like machine translation~\cite{gu-etal-2018-meta} and relation 
classification~\cite{obamuyide-vlachos-2019-model}.
In this work, we extend the conventional MAML~\cite{finn2017model} algorithm, which was initially 
proposed for few-shot learning, as a tool to inject inductive bias, inspired by~\citet{li2018maml, wang2020meta}. For compositional generalization, \citet{lake2019compositional} proposes a meta-learning procedure to train a memory-augmented neural model. However, its meta-learning algorithm is specialized for the SCAN dataset~\cite{lake2018generalization} and not suitable to more realistic datasets.



%% file: sections/conclusion.tex
\section{Conclusion}

Our work highlights the importance of training objectives that select for robust generalization strategies. The meta-learning augmented approach to supervised learning used here allows for the specification of different constraints on learning through the design of the meta-tasks. Our similarity-driven task design improved on baseline performance on two different compositional generalization datasets, by inhibiting the model's ability to memorize large sections of its input. Importantly though the overall approach used here is model agnostic, with portions of it (Str-MAML, Lev-MAML, and Uni-MAML) proving dataset agnostic as well requiring only that the input be a natural language sentence. Our methods are simple to implement compared with other approaches to improving compositional generalization, and we look forward to their use in combination with other techniques to further improve models' compositional ability.

%% file: sections/appendix.tex
\section{Experiments}

\subsection{Details of Base Parsers}

We implemented all models with Pytorch~\cite{paszke2019pytorch}. 
For the LSTM parsers, we use a two-layer encoder and one-layer decoder with attention~\cite{bahdanau2014neural} and input-feeding~\cite{luong-etal-2015-effective}.
We only test bidirectional LSTM encoders, as unidirectional LSTM models do not perform very well in our preliminary experiments. 
For Transformer parsers, we use 2 encoder and decoder layers, 4 attention heads, and a feed-forward dimension of 1024. The hidden size for both LSTM and Transformer models are 256. 
The hyparameters of base parsers are mostly 
borrowed from related work and not tuned, as the primary goal of this work is the MAML training algorithm.
To experiment with a wide variety of possible Seq2Seq models, we also try a Transformer encoder + LSTM decoder and find that this variant actually performs slightly better than both vanilla Transformer and LSTM models.
Further exploration of this combination in pursuit of a better neural architecture for compositional generalization might be interesting for future work.


\subsection{Model Selection Protocol}

In our preliminary experiments on COGS, we find almost all the Seq2Seq models achieve $> 99\%$ in accuracy on the original Dev set. However, their performance on the Gen set diverge dramatically, ranging from $10\%$ to $70\%$. The lack of an informative Dev set makes model selection extremely difficult and difficult to reproduce. This issue might also be one of the factors that results in the large variance of performance reported in previous work.
Meanwhile, we found that some random seeds~\footnote{Random seeds control the initialization of parameters and the order of training batches.} yield consistently better performance than others across different conditions. For example, among the ten random seeds used for Lev-MAML + Transformer on COGS, the best performing seed obtains $73\%$ whereas the lowest performing seed obtains $54\%$.  Thus, it is important to compare different models using the same set of random seeds, and not to tune the random seeds in any model. To alleviate these two concerns, we choose the protocol that is mentioned in the main paper. This protocol helps to make the results reported in our paper reproducible. 

\subsection{Details of Training and Evaluation}

Following~\citet{kim_cogs_2020},
we train all models from scratch using randomly initialized embeddings.
For SCAN, models are trained for 1,000 steps with batch size 128.
We choose model checkpoints based on their performance on the Dev set.
For COGS, models are trained for 6,000 steps with batch size of 128. 
We choose the meta-train learning rate $\alpha$ in Equation~\ref{eq:meta_train}, temperature $\eta$ in Equation~\ref{eq:rel_dist} based on the performance on the Gen Dev set. Finally we use the chosen $\alpha$, $\eta$ to train models with new random seeds, and only the last checkpoints (at step 6,000) are used for evaluation on the Test and Gen set.

\subsection{Other Splits of SCAN}
The SCAN dataset contains many splits, such as Add-Jump, Around Right, and Length split, each assessing a particular case of compositional generalization. We think that MCD splits are more representative of compositional generalization due to the nature of the principle of maximum compound divergence. Moreover, it is more challenging than other splits (except the Length split) according to \citet{furrer2020compositional}. That GECA, which obtains $82\%$ in accuracy on JUMP and Around Right splits, only obtains $< 52\%$ in accuracy on MCD splits in our experiments confirms that MCD splits are more challenging.

 \input{tables/discussion_kernel_analysis}
\input{tables/methods_kernel_examples_2}
\subsection{Kernel Analysis}
 The primary difference between the tree-kernel and string-kernel methods is in the diversity of the examples they select for the meta-test task. The tree kernel selects a broader range of lengths, often including atomic examples, a single word in length, matching a word in the original example from meta-train (see table~\ref{table:discussion-kernels}). By design the partial tree kernel will always assign a non-zero value to an example that is an atom contained in the original sentence. We believe the diversity of the sentences selected by the tree kernel accounts for the superior performance of Tree-MAML compared with the other MAML conditions. The selection of a variety of lengths for meta-test constrains model updates on the meta-train task such that they must also accommodate the diverse and often atomic examples selected for meta-test. This constraint would seem to better inhibit memorizing large spans of the input unlikely to be present in meta-test.

\subsection{Meta-Test Examples}
In Table~\ref{table:kernel-examples-appendix}, we show top scoring examples retrieved by the similarity metrics for two sentences. We found that in some cases (e.g., the right part of Table~\ref{table:kernel-examples-appendix}), the tree-kernel can retrieve examples that diverge in length but are still semantically relevant. In contrast, string-based similarity metrics, especially LevDistance, tends to choose examples with similar lengths.

\subsection{COGS Subtask Analysis}

\begin{figure}
    \centering
\includegraphics[scale=0.7]{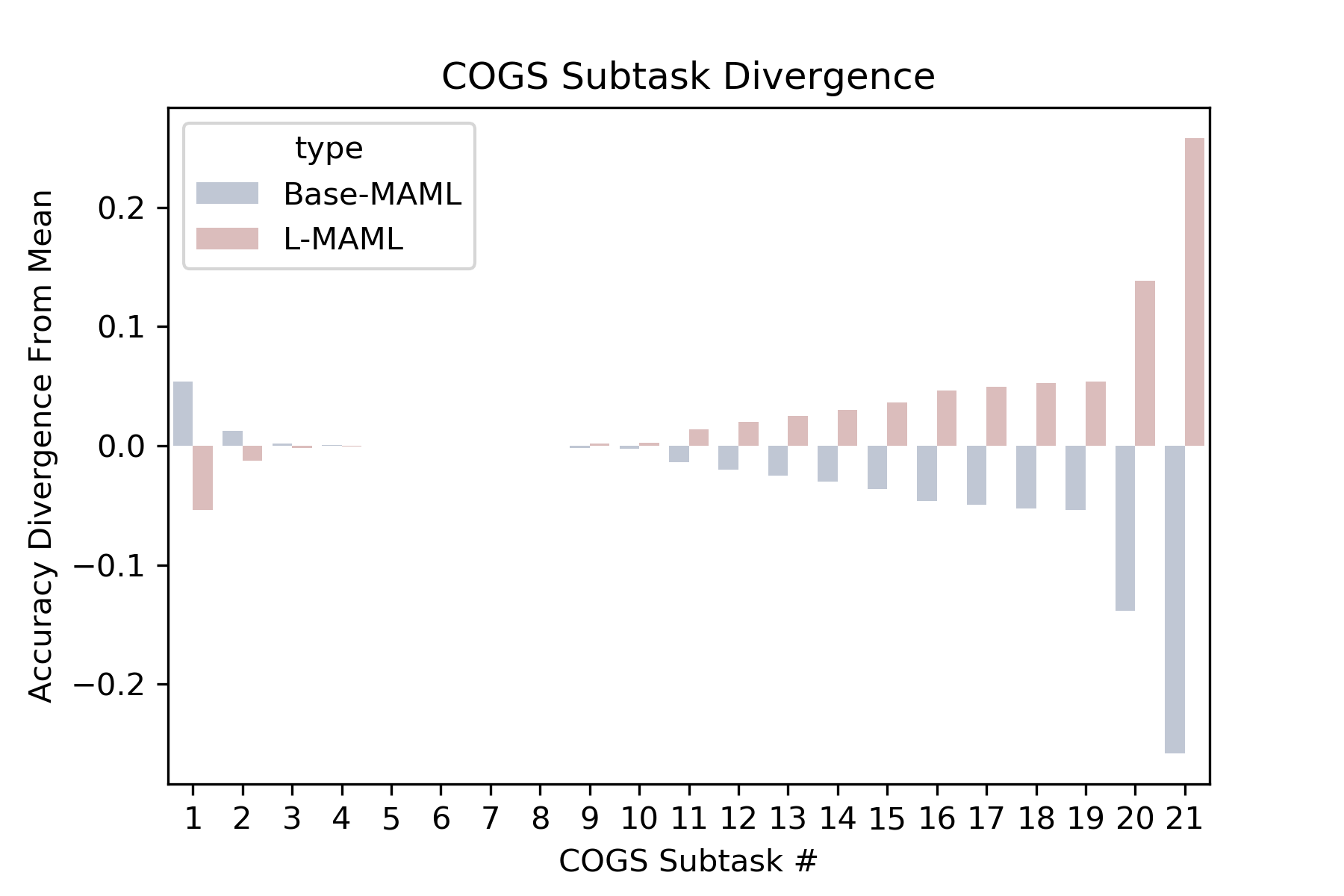}
    \caption{Performance for the Uni-MAML and Lev-MAML conditions compared to the mean of those two conditions. 
    } 
    \label{fig:Cogssubtask}
\end{figure}

We notice distinct performance for different conditions on the different subtasks from the COGS dataset. In Figure \ref{fig:Cogssubtask} we show the performance of the Uni-MAML and Str-MAML conditions compared with the mean of those conditions. Where the bars are equal to zero the models' performance on that task is roughly equal. \\

\noindent Full task names for figure \ref{fig:Cogssubtask}:\\
(1) prim$\rightarrow$subj proper, \\
(2) active$\rightarrow$passive,\\
(3) only seen as unacc subj $\rightarrow$ unerg subj,\\
(4) subj$\rightarrow$obj proper,\\
(5) only seen as unacc subj $\rightarrow$ obj omitted transitive subj,\\
(6) pp recursion,\\
(7) cp recursion,\\
(8) obj pp$\rightarrow$subj pp,\\
(9) obj$\rightarrow$subj common,\\
(10) do dative$\rightarrow$pp dative,\\
(11) passive$\rightarrow$active,\\
(12) only seen as transitive subj $\rightarrow$ unacc subj,\\
(13) obj omitted transitive$\rightarrow$transitive, \\
(14) subj$\rightarrow$obj common, \\
(15) prim$\rightarrow$obj proper, \\
(16) obj$\rightarrow$subj proper, \\
(17) pp dative$\rightarrow$do dative, \\
(18) unacc$\rightarrow$transitive, \\
(19) prim$\rightarrow$subj common, \\
(20) prim$\rightarrow$obj common, \\
(21) prim$\rightarrow$inf arg.

%% file: tables/discussion_kernel_analysis.tex
\begin{table*}
\centering
\resizebox{2.0 \columnwidth}{!}{%
\begin{tabular}{ cc } 
    \begin{tabular}{lrrr}
    \textbf{Partial Tree Kernel}             &  \textbf{top 10} & \textbf{100} & \textbf{1000}  \\
    \toprule
    Mean Example Length (chars) & 26.71 & 26.59 & 29.87 \\
    Std dev & $\pm$ 6.80 & $\pm$ 7.61  & $\pm$ 8.85 \\
    \hline
    Mean No. of Atoms & 0.46 & 0.81 & 1.13\\
    Std dev & $\pm$ 0.67 & $\pm$ 1.05 & $\pm$ 0.81\\
    \bottomrule
    \end{tabular}

    \begin{tabular}{lrrr}
    \textbf{LevDistance}             &  \textbf{top 10} & \textbf{100} & \textbf{1000}  \\
    \toprule
    Mean Example Length (chars) & 31.04 &  30.45  & 29.28 \\
    Std dev & $\pm$ 2.80 &  $\pm$ 3.77 & $\pm$ 4.78 \\
    \hline
    Mean No. of Atoms & 0.00 & 0.00 & 0.02\\
    Std dev & $\pm$ 0.00 & $\pm$ 0.02 & $\pm$ 0.17\\
    \bottomrule
    \end{tabular}

\end{tabular}
}
\caption{Analyses of kernel diversity. Reporting mean example length and number of atoms for the top k highest scoring examples for each kernel. Note that atoms are only counted that also occur in the original example.}
\label{table:discussion-kernels}
\end{table*}

%% file: tables/methods_kernel_examples_2.tex
\begin{table*}[t!]
  \centering
    \resizebox{2.0\columnwidth}{!}{ 
\begin{tabular}{cc}
      \begin{tabular}{lr}
    \multicolumn{2}{c}{\textbf{Source Example}: Emma lended the donut to the dog .} \\
      \\
      \toprule
      \emph{Neighbours using Tree Kernel} & Similarity\\
      \hline
Emma was lended the donut . & 0.74 \\
The donut was lended to Emma . & 0.62 \\
Emma lended the donut to a dog . & 0.55 \\
Emma lended Liam the donut . & 0.55 \\
Emma lended a girl the donut . & 0.55 \\
       \\
        \emph{Neighbours using String Kernel}\\
       \hline
Emma lended the donut to a dog . & 0.61 \\
Emma lended the box to a dog . & 0.36 \\
Emma gave the cake to the dog . & 0.33 \\
Emma lended the cake to the girl . & 0.33 \\
Emma lended the liver to the girl . & 0.33 \\
      \\
      \emph{Neighbours using LevDistance} & \\
      \hline
Emma lended the donut to a dog . & -1.00 \\
Emma loaned the donut to the teacher . & -2.00 \\
Emma forwarded the donut to the monster . & -2.00 \\
Emma gave the cake to the dog . & -2.00 \\
Charlotte lended the donut to the fish . & -2.00 \\
      \bottomrule
      \end{tabular}
    &
      \begin{tabular}{lr}
    \multicolumn{2}{c}{\textbf{Source Example}:  The crocodile valued that a girl snapped .} \\
      \\
      \toprule
      \emph{Neighbours using Tree Kernel} & Similarity\\
      \hline
A girl snapped . & 0.55 \\
A rose was snapped by a girl . & 0.39 \\
The cookie was snapped by a girl . & 0.39 \\
girl & 0.32 \\
value & 0.32 \\
       \\
        \emph{Neighbours using String Kernel}\\
       \hline
The crocodile liked a girl . & 0.28 \\
The girl snapped . & 0.27 \\
The crocodile hoped that a boy observed a girl . & 0.26 \\
The boy hoped that a girl juggled . & 0.15 \\
The cat hoped that a girl sketched . & 0.15 \\
      \\
      \emph{Neighbours using LevDistance} & \\
      \hline
The crocodile liked a girl . & -3.00 \\
The boy hoped that a girl juggled . & -3.00 \\
The cat hoped that a girl sketched . & -3.00 \\
The cat hoped that a girl smiled . & -3.00 \\
Emma liked that a girl saw . & -4.00 \\
      \bottomrule
      \end{tabular}
\end{tabular}
  }
  \caption{Top scoring examples according to the tree kernel, string kernel and Levenshtein distance for two sentences and accompanying scores.}
  \label{table:kernel-examples-appendix}
  \end{table*}